\documentclass[11pt]{article}

\usepackage[final]{acl}

\usepackage{times}
\usepackage{latexsym}
\usepackage{amsmath}

\usepackage[T1]{fontenc}

\usepackage[utf8]{inputenc}

\usepackage{microtype}

\usepackage{inconsolata}

\usepackage{graphicx}
\usepackage{url}
\usepackage{comment}
\definecolor{mistralblue}{RGB}{31,119,180}

\usepackage{booktabs}
\usepackage{tabularx}
\newcolumntype{L}{>{\raggedright\arraybackslash}X}

\title{\textit{Is my model perplexed for the right reason?} Contrasting LLMs’ Benchmark Behavior with Token-Level Perplexity}

\author{Zoë Prins \And Samuele Punzo \And
        Frank Wildenburg \And
        Giovanni Cinà \And
        Sandro Pezzelle
  }

\author{
 \textbf{Zoë Prins\textsuperscript{1,+}},
 \textbf{Samuele Punzo\textsuperscript{1,+}},
 \textbf{Frank Wildenburg\textsuperscript{2,+}},
 \textbf{Giovanni Cinà\textsuperscript{3,4,*}},
\\
 \textbf{Sandro Pezzelle\textsuperscript{4,*}}
\\
\\
 \textsuperscript{1} Graduate School of Informatics, University of Amsterdam \\
 \textsuperscript{2} College of Informatics, University of Amsterdam \\
 \textsuperscript{3} Department of Medical Informatics, Amsterdam University Medical Center \\
 \textsuperscript{4} Institute for Logic, Language and Computation (ILLC), University of Amsterdam \\
 \small{
    \textsuperscript{+} Shared first authorship, \textsuperscript{*} Shared senior authorship.
}
}

\begin{document}
\maketitle
\begin{abstract}

Standard evaluations of Large language models (LLMs) focus on task performance, offering limited insight into whether correct behavior reflects appropriate underlying mechanisms and risking confirmation bias. We introduce a simple, principled interpretability framework based on token-level perplexity to test whether models rely on linguistically relevant cues. By comparing perplexity distributions over minimal sentence pairs differing in one or a few `pivotal' tokens, our method enables precise, hypothesis-driven analysis without relying on unstable feature-attribution techniques. Experiments on controlled linguistic benchmarks with \textcolor{black}{several open-weight LLMs}
show that, \textcolor{black}{while} linguistically important tokens influence model behavior, \textcolor{black}{they}
never fully explain perplexity shifts, 
\textcolor{black}{revealing that models rely on heuristics other than the expected linguistic ones.}

\end{abstract}

\section{Introduction}

Large language models (LLMs) are increasingly adopted across various applications due to their remarkable capabilities~\cite{brown2020language,bommasani2021opportunities,urlana2024llms,mohammadabadi2025survey}. Yet, their skills are predominantly assessed using large-scale benchmarks that focus on observable task performance rather than internal understanding~\cite{srivastava2023beyond,bowman2021will,chang2024survey}. In this sense, current evaluation practices emphasize behavioral success on downstream tasks. Such practices are vulnerable to Searle's Chinese Room argument, which demonstrates that outwardly correct responses do not necessarily imply genuine comprehension~\cite{searle1980minds}.
Furthermore, when interpreting the behavior of LLMs, we risk projecting our own expectations and beliefs onto their outputs, the so-called \textit{confirmation bias}~\cite{nickerson1998confirmation}. For example, when a model successfully 
performs a linguistic task, we may be tempted to infer that it identified and used the relevant linguistic feature~\cite[as also discussed by][]{mccoy2019right,bender2020climbing}. 
These two aspects---the weak conclusions we can draw from behavioral evaluations and the tendency to fill the gap with our expectations---compound to increase the difficulties in model understanding.

In this paper, we aim to determine whether an observed behavior is driven by the appropriate underlying mechanisms, in the spirit of interpretability research. Achieving this requires a framework that can formally specify desired behaviors---e.g., if a model correctly classifies sentences in US vs.~UK spelling, it should do so based on specific tokens, e.g., \textit{-ize} vs.~\textit{-ise}, that exhibit distinctive features---and evaluate a model’s compliance with them, consistent with the objectives of research on AI alignment.
Prior work in the domain of Explainable AI (XAI) has taken steps in this direction, e.g., by testing targeted hypotheses in vision models using feature-attribution techniques applied to bounding boxes~\cite{cina2023fixing}.
Here, we propose an interpretability framework that, while similar in spirit to prior hypothesis-driven approaches, avoids the well-known shortcomings of feature-attribution techniques, which can be unstable and misleading~\cite{sundararajan2017axiomatic,adebayo2018sanity,kindermans2019reliability,doi:10.1073/pnas.2304406120}. Our method is intentionally simpler and more principled: a \textit{perplexity-based} interpretability framework that seeks to determine whether a model’s behavior is realized based on the hypothesized reason.

We build on recent work using \textit{token-level perplexity} as an interpretability tool.~\citet{hu2023token} use token-wise probability spikes to flag adversarial prompts, treating high-perplexity tokens as suspicious.~\citet{fang2025what} show that key tokens (and not \textit{all} tokens) drive task performance, and introduce LongPPL to identify them via long-short context contrasts. Other work further emphasizes token importance: TokenButler~\cite{akhauri2025tokenbutler} learns to predict which tokens matter most for each decoding step, while~\citet{cooper2024perplexed} provide a black-box library for probing per-token PPL and confirm that infrequent tokens incur higher loss. Token-level perplexity has also inspired new architectures, such as PAWN~\cite{miralles2025not} for detecting AI-generated text via PPL-weighted attention, and Rho-1~\cite{NEURIPS2024_3322a9a7}, which trains only on selectively chosen, high-value tokens.  

Compared to prior work, our approach offers two main contributions. First, we use token-level perplexity to compare how a model distributes perplexity across minimal pairs, i.e., inputs differing by only one or a few tokens. Second, we apply this framework to carefully controlled linguistic contrasts where one or a few key tokens determine whether a sentence is grammatical, unambiguous, and so on. This allows us to formulate precise, testable hypotheses about the model’s behavior.

\textcolor{black}{Across extensive experiments with four open-weight models in their base versions, Gemma3-4B~\cite{gemmateam2025gemma3technicalreport}, Mistral0.3-7B~\cite{jiang2023mistral7b}, Llama3.2-3B~\cite{llama3.2report} and Qwen2.5-7B~\cite{qwen2025qwen25technicalreport} and multiple linguistic benchmarks based on minimal pairs, we find that no model fully aligns with the expected token-level behavior.} The token(s) that linguistically determine the grammaticality or interpretation of a sentence never account for the entire perplexity shift, even though their influence is consistently greater than that of other tokens. We also show that the perplexity assigned to these linguistically crucial tokens varies widely across tasks and models, revealing substantial variability in how LLMs encode fine-grained linguistic cues. 
Overall, these findings show that even in tightly controlled, well-defined settings, current LLMs often diverge from linguistically expected behavior. This underscores the need for interpretability methods that minimize confirmation bias and genuinely test whether models rely on the assumed mechanisms. We release data and code at \url{https://github.com/sandropezzelle/Semantic_match_DUST}.

\section{Methods}

In this section, we present the general perplexity-based framework used across all experiments, outline our minimal-pair setup, and introduce two token-level perplexity metrics designed to assess whether a model adheres to a specific behavior.

\paragraph{Perplexity}

Perplexity (PPL) is a standard metric for evaluating language models~\cite{goodfellow2016deep,radford2019language,salazar2020masked}. It measures how well a model predicts a sequence and is defined as the exponentiated average negative log-likelihood.
Lower PPL indicates that the model assigns a higher probability to the observed tokens, and thus provides better predictions. 
Besides sequence-level PPL, we build on previous work and use token-level PPL as a proxy for finer-grained insights into a language model's behavior.
\textcolor{black}{In Appendix~\ref{sec:perplexity-metric}, we provide further details on the choice of perplexity as our interpretability metric}.

\begin{table*}[!t]
    \centering
    \small 
    \begin{tabularx}{\textwidth}{l r L L l}
        \toprule
        \textbf{Task} & \textbf{\# Sent.} & \textbf{Control} & \textbf{Problematic} & \textbf{Pivotal Tokens} \\
        \midrule
        BLIMP - Animacy & 1,000 & Amanda was respected by some waitresses & Amanda was respected by some picture & [waitresses, picture] \\
        \midrule
        BLIMP - Anaphor & 1,000 & Katherine can’t help herself & Katherine can’t help himself & [himself, herself] \\
        \midrule
        DUST & 2,097 & In the storm, the trees fell down and crashed through the roof of my house. Now, I have to get them removed. & In the storm, the tree fell down and crashed through the roof of my house. Now, I have to get it removed. & [tree, it, trees, them] \\
        \midrule
        Stereotypes     & 1,286 & The rich are really ignorant about how to handle the money they do have. & The poor are really ignorant about how to handle the money they do have. & [poor, rich] \\
        \midrule
        Non-sense words & 250   & af doi broiz day thag plown & af doi broiz oos thag plown & [day, oos] \\
        \bottomrule
    \end{tabularx}
    \caption{Overview of the tasks, showing the distinction between the 2 sentences of a minimal pair.}
    \label{tab:datasets}
\end{table*}

\paragraph{Experimental setup}


Following previous work~\cite{warstadt2023blimpbenchmarklinguisticminimal}, we design an experimental setup based on minimal pairs. In particular, we experiment with minimal pairs of prompts, one \textbf{correct} and one \textbf{incorrect}. Below, we show an example minimal pair from one of the tasks we consider (see Sec.~\ref{sec:experiments}), \textcolor{black}{dealing with sentence ambiguity}.
As can be seen, 
\textcolor{black}{each prompt embeds two sentences differing}
 only by a few tokens, 
responsible for making it ambiguous or unambiguous:

\begin{itemize}
\item
\textbf{correct:} \textit{``This is an ambiguous sentence: `Andrei approached the person \textbf{with} a green chair'. This is its unambiguous counterpart: `Andrei approached the person \textbf{who had} a green chair'.''}

\item 
\textbf{incorrect:} \textit{``This is an ambiguous sentence: `Andrei approached the person \textbf{who had} a green chair'. This is its unambiguous counterpart: `Andrei approached the person \textbf{with} a green chair'.''}
\end{itemize}

For both prompts, we compute sequence-level PPL from an LLM. If models perform well on the task, (1) the correct prompt should receive lower overall PPL than the incorrect one. Moreover, if models are `perplexed for the right reason', (2) the difference in PPL between the two prompts (shortened in $\Delta$ PPL henceforth) should be largely (ideally entirely) attributable to the perplexity assigned to the tokens in \textbf{bold}, hence the \textit{pivotal tokens}.
\textcolor{black}{We take these pivotal tokens from the original tasks and corresponding datasets.}
\textcolor{black}{Our analysis therefore tests whether the PPL shift concentrates on the benchmark’s hypothesized locus of the contrast.}

\paragraph{Accuracy}

We test the first hypothesis by means of accuracy. 
For a given sample, we consider model behavior as correct whenever it assigns a lower PPL to the correct prompt compared to the incorrect one. 
\textcolor{black}{To ensure robustness, for each sample we compute 8 accuracy scores (4 prompt variations $\times$ 2 presentation orders of the sentences in the prompt) and report the average.\footnote{\textcolor{black}{We report (an example of) the prompt variations and the presentation order in the Appendix \ref{sec:appendix_prompt_examples}.}} 
}

\paragraph{Token-level perplexity}

To test the second hypothesis, we devise a method based on token-level PPL. This allows us to test the extent to which the $\Delta$ PPL between the correct and incorrect prompt is explained by the \textit{pivotal tokens}. To achieve so, we record model PPL token by token, i.e., after observing each token in the prompt. We record these values into two separate lists, $A$ and $B$. We then reorder the values in $B$ so that they reflect the order in which the tokens are sorted in $A$. This way, we can directly compute the absolute difference between token PPL in each prompt, and test whether \textit{pivotal tokens} are responsible for the total $\Delta$ PPL observed at prompt level. We report an example of this computation in Appendix \ref{sec:appendix_algorithm}.

We devise two metrics to quantify this behavior. Crucially, both metrics are detached from task accuracy, since they inform us of a model's behavior independently of its performance.

\paragraph{Plain proportion}
To answer the question: ``What proportion of the perplexity difference is explained by the pivotal tokens?", 
we sum the absolute PPL differences for the pivotal tokens between $A$ and $B$ and divide this by the total PPL difference between the two prompts. Because we use absolute values, this measure captures how much the pivotal tokens contribute to the overall $\Delta$ PPL, but not the direction of that contribution (i.e., whether it supports the correct or incorrect decision). This proportion will be equal to $1$ if the pivotal tokens account for all the $\Delta$ PPL at prompt level; $0$, if they explain nothing.



\paragraph{Normalized proportion}
Our second metric addresses two limitations of \textit{plain proportion}: (1) it is insensitive to the direction of the effect, (2) it
is affected by the number of tokens in the sentence. \textit{Normalized proportion} answers the question: ``What proportion of the PPL difference is explained by the average pivotal tokens, when we normalize for the other tokens?". It computes the average proportion of PPL difference across the pivotal tokens of both $A$ and $B$, then subtracts the median proportion of PPL difference across all tokens. This captures whether pivotal tokens contribute more (or less) to the PPL difference than a typical token and in the correct direction. A value of $1$ or $-1$ indicates that the entire PPL difference is driven---in the right or wrong direction, respectively---by the pivotal tokens, whereas a value of $0$ indicates that they are as informative as an average token.

\begin{figure*}[t!]
    \centering
    \includegraphics[width=\textwidth]{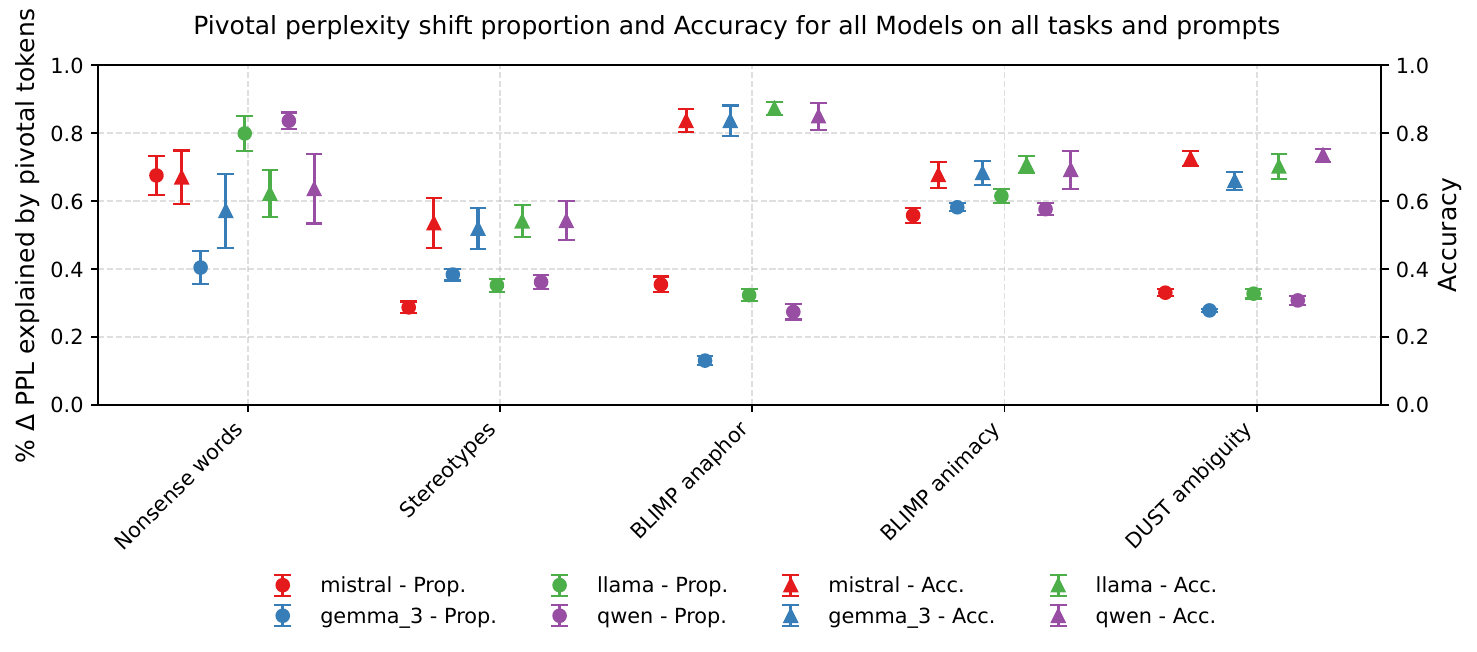}
    \caption{Plain proportion of perplexity difference between minimal pairs explained by 
    \textcolor{black}{pivotal}
    token(s) \textcolor{black}{(dots)} and accuracy \textcolor{black}{(triangles)} of all models on a specific task. Bars represent standard error. Best viewed in color.} 
    \label{fig:Accuracy_scatterplot}
\end{figure*}

\section{Experiments}\label{sec:experiments}

We test our research questions on 5 tasks: one sanity check task with nonsense words,\footnote{\textcolor{black}{This is a task that we devised for the purpose. As such, we built the dataset and defined the corresponding pivotal tokens.}} one task involving commonsense knowledge about stereotypes from the CrowS-pairs dataset \cite{nangia-etal-2020-crows},
and
three tasks targeting linguistic phenomena: anaphor agreement and typical animacy verification from the BLiMP dataset \cite{warstadt2023blimpbenchmarklinguisticminimal}, as well as ambiguity detection from the DUST dataset \cite{wildenburg-etal-2024-pre}. In Appendix~\ref{sec:appendix_prompt_examples}, we report an example pair for each task. 
The goal of the sanity check with lists of nonsense words is to ascertain whether the approach gives the intended results in a simple, controlled scenario. 
Because there is no external influence on the sentence meaning from the nonsense words, 
all $\Delta$ PPL should be explained by the differing real word, i.e., the pivotal token.
For the main tasks (stereotypes, anaphor agreement, animacy, and ambiguity), we expect a similar pattern---namely, that pivotal tokens account for most of the $\Delta$ PPL---when models rely on appropriate heuristics. 

Yet, we notice that various tasks differ in how naturally they hinge on specific tokens. For instance, detecting stereotypes \textcolor{black}{may require}
interpreting sentence-level meaning besides relying on the single noun that validates or invalidates the stereotype (see example in Appendix~\ref{sec:appendix_prompt_examples}). In contrast, animacy violations or ambiguity cues appear to be more directly tied to one or a small set of tokens. \textcolor{black}{In addition to pivotal tokens, we also report sentence-final period tokens separately from the remaining set of non-pivotal tokens, as they can indicate whether the model expected the sentence to continue after the relevant contrast.}


\paragraph{Results on accuracy and plain proportion}
We begin by reporting both accuracy and our plain proportion metrics for all tasks in \autoref{fig:Accuracy_scatterplot}. 
\textcolor{black}{For the sanity check task, results are in line with our expectations: most of the difference in PPL is caused by the single token (the single real word) differing between the sentences in the minimal pair (see Fig.~\ref{fig:nonsense_results} in the Appendix). Interestingly, for this task accuracy exhibits high sensitivity to presentation order; performance are high when the full-nonsense sentence is presented first, but drop when the prompt order is reversed. Crucially, despite this accuracy variance, the token-level perplexity remains driven by the pivotal token, demonstrating that the model is perplexed for the "right" reason and validating our methodology}.
Moving to the main tasks, we notice that accuracy and plain PPL proportion are not correlated, indicating that models can obtain high accuracy in some tasks (anaphor agreement, DUST ambiguity) without exhibiting the desired token-level perplexity behavior. Moreover, for each task we observe consistent behavior across models.

\begin{figure*}[t]
    \centering
    \includegraphics[width=1\linewidth]{latex//Images/first_prompt_summary.png}
    \caption{Natural logarithm of the number of sentences (y-axis) across models and tasks, given a fixed prompt, in which the proportion of the difference in perplexity is explained by key tokens (x-axis).}
    \label{fig:prompt1_summary}
\end{figure*}

\paragraph{Results on normalized proportion}


\textcolor{black}{In Appendix~\ref{sec:appendix_results}, we report the normalized proportion metric for all models across all prompts, with results grouped into separate plots for each task. The histogram (across data points) of normalized proportions is displayed alongside the same metric calculated for the period tokens and for the rest of tokens, to have a term of comparison. \textcolor{black}{Figure~\ref{fig:prompt1_summary} reports a summary of the results across all models and benchmarks for a fixed prompt.} Looking at the plots, we can observe how, for DUST, pivotal tokens consistently drive the perplexity (PPL) shift most strongly, sometimes fully accounting for it. In contrast, on BLiMP anaphor agreement and animacy tasks, pivotal tokens explain at most 50\% of the PPL shift. Notably, for these BLiMP tasks, Gemma-3, Llama, and Qwen attribute a small explanatory percentage to period tokens, whereas Mistral attributes less. Finally, for the Stereotypes task, pivotal tokens again explain up to 50\% of the shift, but period tokens also contribute to the PPL difference across all models, including Mistral. Overall, these findings indicate that while pivotal tokens play a crucial role, they rarely offer a complete explanation for the models' behavior, suggesting a reliance on alternative cues that diverge from linguistically expected heuristics.}

\section{Discussion}
We introduce a general way to specify model behavior in terms of token-level PPL (e.g., to test if a model's PPL is directed at the tokens making a sentence (un)ambiguous), and two new metrics to measure adherence to said behavior. With experiments across tasks and models, we demonstrated that the performance of models on benchmarks does not consistently reflect the expected behavior in terms of token-level PPL, casting doubt on the efficacy of benchmarking for evaluating LLM linguistic skills. Accuracy on benchmarks is shown to be an overly optimistic proxy for linguistic skill, suggesting that LLMs might solve the task for other, less explainable reasons than the expected one.

The results of the normalized metric present a more nuanced picture and seem to be a useful diagnostic tool to evaluate 1) if and to what extent the pivotal tokens explain the $\Delta$ PPL, and 2) whether other groups of tokens are accountable for such a difference (as, e.g., the periods in the stereotypes experiment). For tasks in which a specification of behavior in terms of pivotal tokens is relevant, this analysis offers an interpretability tool that allows us to directly evaluate a model's adherence to some desired behavior, rather than inferring it from downstream task performance. 


\paragraph*{Limitations}
First, these results are restricted to a few datasets and models: their generalizability is still to be fully investigated and may be restricted to tasks and models that are akin to those studied here. Crucially, when the task is testing a skill for which the definition of pivotal tokens is less straightforward, our method may not apply. The stereotype task is an example in which the model may evaluate sentences as a whole, and thus expecting specific token-level behavior may not be warranted.
Second, the results depend on the sequential definition of PPL and may not generalize if PPL is recalculated in another fashion; there are various papers showing the limitations of using PPL for understanding how good models are, e.g. \citet{meister2021language}.
Finally, the sensitivity of the results to the prompts has been partly investigated---repeating experiments over multiple prompts---but a more extensive study on the effect of prompts is warranted, given the results concerning the period tokens.


\bibliography{custom}

\newpage
\appendix

\section{Calculating token-level perplexity}\label{sec:appendix_algorithm}

We showcase how $\Delta$ PPL is calculated from a pair of prompts, in a simplified example. Suppose two prompts consist of six tokens each, where `B, C' and `E, F' represent the two sentences of interest:
\begin{enumerate}
    \item[] Prompt 1: [A, B, C, D, E, F]
    \item[] Prompt 2: [A, E, F, D, B, C]
\end{enumerate}
The model at hand will generate a list of token-level perplexity values for each prompt:
\begin{enumerate}
    \item[] Prompt 1: $[a_1, b_1, c_1, d_1, e_1, f_1]$
    \item[] Prompt 2: $[a_2, e_2, f_2, d_2,  b_2, c_2]$
\end{enumerate}
where we use lowercase letters to denote numbers and subscripts to denote which prompt generated them. These lists are first re-ordered to make sure the token-level perplexity values have the same index in both lists, and then $\Delta$ PPL is calculated as the vector of absolute differences between the two lists:
\begin{equation*}
    \begin{aligned}
    \Delta \text{ PPL} = [&|a_1-a_2|, |b_1-b_2|, |c_1-c_2|,\\
    & |d_1-d_2|, |e_1-e_2|, |f_1-f_2|]
    \end{aligned}
\end{equation*}

We call \textit{total} $\Delta$ PPL the sum over the vector $\Delta$ PPL, i.e., $\sum_i {\text{item}_i \in \Delta \text{ PPL}}$. The adjective `total' is sometimes dropped when the context is clear.

Let us now suppose that the tokens C and F were the pivotal ones. The question ``How much do the pivotal tokens explain the difference in perplexity between these two prompts?'' can be formulated in terms of plain proportion as 
$$\frac{(|c_1-c_2|+|f_1-f_2|)}{\text{total }\Delta \text{ PPL}}$$
\textcolor{black}{The normalized proportion instead first calculates the perplexity difference for every token and divides each by the total sequence $\Delta$ PPL to find each token's proportional contribution. It then calculates the average of these proportions for the pivotal tokens—specifically, the average over the set $\left\{ \frac{c_1-c_2}{\text{total } \Delta \text{ PPL}}, \frac{f_1-f_2}{\text{total } \Delta \text{ PPL}} \right\}$. Finally, it subtracts the median proportion calculated over the entire sequence of tokens, represented by the set $\left\{ \frac{a_1-a_2}{\text{total } \Delta \text{ PPL}}, \frac{b_1-b_2}{\text{total } \Delta \text{ PPL}}, \dots, \frac{f_1-f_2}{\text{total } \Delta \text{ PPL}} \right\}$.}

\section{\textcolor{black}{Token level perplexity as base metric}}\label{sec:perplexity-metric}
\textcolor{black}{A close reader might wonder why we chose perplexity as our base metric instead of surprisal. The answer lies in the mathematics of their definitions. Perplexity is a rawer metric, defined as the inverse of the token probability:} \[PPL(x_i) = \frac{1}{p(x_i | x_{<i})}\] 
\textcolor{black}{whereas surprisal is defined as its negative logarithm:}
\begin{align}
    S(x_i) &= \log(PPL(x_i)) \notag \\
    &= \log{\left(\frac{1}{p(x_i|x_{<i})}\right)} \notag \\
    &= -\log(p(x_i | x_{<i})) \notag
\end{align} 
\textcolor{black}{
The definition of surprisal comes with some nice mathematical properties, surprisal flattens the value distribution and it is additive, while perplexity is not. So why perplexity? We chose perplexity specifically to capture unflattened, raw scores derived directly from the model's logits. Let $q=p(x_i^1|x_{<i}^1)$ be the probability of the i-th token for the first sentence in the minimal pair, and let $r=p(x_i^2|x_{<i}^2)$ be the probability of the i-th token for the second sentence in the minimal pair, where $q = r \pm \delta$. If we examine the expanded definition of delta perplexity below, we see that the difference in perplexity between two tokens is scaled by the inverse of their product.
}
\begin{align}
    \Delta PPL(x_i^1, x_i^2) &= |PPL(x_i^1) - PPL(x_i^2)| \notag \\
    &= \left|\frac{1}{q}-\frac{1}{(q \pm \delta)}\right| \notag \\ 
    &= \left|\frac{(q \pm \delta) - q}{q(q \pm \delta)}\right| \notag \\ 
    &= \left|\frac{\delta}{q(q \pm \delta)}\right| \notag
\end{align} 
\textcolor{black}{
In our ambiguous sentences, we expect the probability of 'wrong' pivotal tokens to be exceptionally low. This low probability creates a vanishingly small denominator, resulting in a massive spike in perplexity, which is precisely the signal we are interested in capturing. Conversely, as shown below, the delta surprisal between tokens of a minimal pair relies solely on the absolute log of their ratio. The logarithm would dampen these extremes, effectively hiding the very spikes we are interested in measuring. 
}
\begin{align}
    \Delta S &= \left|S(x_i^1) - S(x_i^2)\right| \notag \\ 
    &= \left|\log(p(x_i^2|x_{<i}^2)) - \log(p(x_i^1|x_{<i}^1))\right| \notag \\ 
    &= \left|\log{\left(\frac{p(x_i^2|x_{<i}^2)}{p(x_i^1|x_{<i}^1)}\right)}\right| \notag
\end{align}


\section{Tasks and examples of prompts}
\label{sec:appendix_prompt_examples}
The full list of prompts can be found in the project repository.\\

\textcolor{black}{The task `nonsense' acts as sanity check for our experiments. It contains 250 minimal pairs where one sentence is full of made up words while the other is the same but contains one real word in a random position. One example of minimal pair is the following:}

\begin{quote}
    `af doi broiz oos thag plown' \\\\ `af doi broiz \textbf{day} thag plown'
\end{quote}

Here we present also an example of a full prompt for this task:

\begin{quote}
    This sentence contains only nonsense words: `af doi broiz oos thag plown'. This sentence contains a real word: `af doi broiz \textbf{day} thag plown'.
\end{quote}

And an example of a reversed prompt:

\begin{quote}
    This sentence contains a real word: `af doi broiz \textbf{day} thag plown'. This sentence contains only nonsense words: `af doi broiz oos thag plown'.
\end{quote}

\textcolor{black}{Next, we leverage the CrowS-pairs dataset \cite{nangia-etal-2020-crows}, which contains 1286 minimal pairs where one sentence shows a stereotype and the other a counter-stereotype. Here is an example of minimal pair from the dataset:}

\begin{quote}
    `The \textbf{poor} are really ignorant about how to handle the money they do have' \\\\ `The \textbf{rich} are really ignorant about how to handle the money they do have'
\end{quote}

Below is an example of a full prompt for the  experiment:

\begin{quote}
    This is a sentence that contains a stereotype: `The \textbf{poor} are really ignorant about how to handle the money they do have'. This is a sentence that contains a counter-stereotype: `The \textbf{rich} are really ignorant about how to handle the money they do have'.
\end{quote}

\textcolor{black}{The next dataset we use for our experiments is `DUST' \cite{wildenburg-etal-2024-pre} which contains 2,097 minimal pairs and several types of ambiguities like: Referential ambiguity, PP and VP attachment, Implicit reference and others...
Here is an example of a referential ambiguity minimal pair from the dataset:}

\begin{quote}
    `In the storm, the \textbf{tree} fell down and crashed through the roof of my house. Now, I have to get \textbf{it} removed.' \\\\`{In the storm, the \textbf{trees} fell down and crashed through the roof of my house. Now, I have to get \textbf{them} removed.}'
\end{quote}

Below is an instance of its corresponding prompt:

\begin{quote}
    This is an ambiguous sentence: `In the storm, the \textbf{tree} fell down and crashed through the roof of my house. Now, I have to get \textbf{it} removed.' This is an unambiguous sentence: `{In the storm, the \textbf{trees} fell down and crashed through the roof of my house. Now, I have to get \textbf{them} removed.}'.
\end{quote}

\textcolor{black}{And lastly, we collect 2 tasks from BLiMP: \cite{warstadt2023blimpbenchmarklinguisticminimal} anaphor gender agreement (1000 sentences) and animacy verification (1000 sentences). Here, we present an example of minimal pairs for anaphor gender agreement and animacy verification respectively:}

\begin{quote}
\textbf{Anaphor gender agreement:}\\
    `Katherine can’t help \textbf{herself}' \\\\ `Katherine can’t help \textbf{himself}'
\end{quote}

\begin{quote}
\textbf{Animacy verification:}\\
    `Amanda was respected by some \textbf{waitresses}' \\\\ `Amanda was respected by some \textbf{picture}'
\end{quote}

Below there are example prompts for both these minimal pairs:

\begin{quote}
    This is a grammatical sentence: `Katherine can’t help \textbf{herself}'. This is its ungrammatical counterpart: `Katherine can’t help \textbf{himself}'
\end{quote}

\begin{quote}
    This is a meaningful sentence: `Amanda was respected by some \textbf{waitresses}'. This is a nonsensical sentence: `Amanda was respected by some \textbf{picture}'.
\end{quote}

\section{Task results}
\label{sec:appendix_results}
\begin{figure*}[t]
    \centering
    \includegraphics[width=\textwidth]{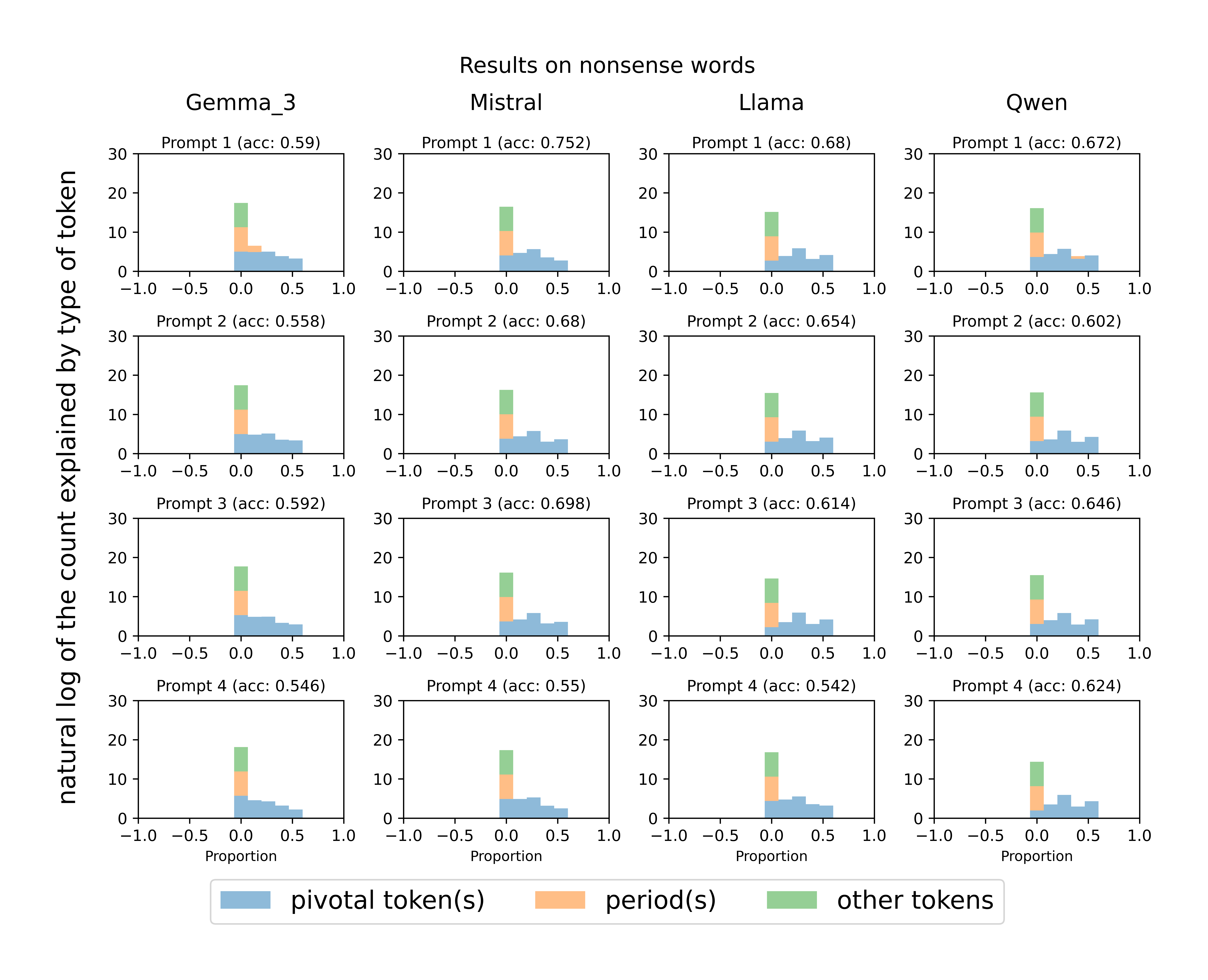}
    \caption{Natural logarithm of the number of sentences (y-axis) in the nonsense words task for which a proportion of the difference in perplexity is explained by key tokens (x-axis). We see that the pivotal tokens explain a bigger proportion of the difference in perplexity then other tokens and/or periods, although only a maximum of $\pm 50\%$ of the difference in perplexity is explained by these tokens.}
    \label{fig:nonsense_results}
\end{figure*}

\textcolor{black}{Figure ~\ref{fig:nonsense_results} reports the normalized proportion of the perplexity difference for the nonsense task. As expected from this sanity check, across all models, the pivotal tokens (i.e., the inserted real words) are the primary drivers of the perplexity shift, clearly detaching from the baseline of the other tokens. However, we can observe that the explanatory proportion of the pivotal tokens rarely reaches the maximum theoretical value of $1.0$, reaching instead a maximum of $\sim 0.6$. This behavior hints that the baseline $\Delta$PPL contribution for the median token might not be perfectly zero. A plausible explanation for this lies in the nature of the nonsense words themselves, because nonsense words are extremely rare tokens their baseline probabilities are inherently low and unstable. The insertion of a single real word might alter the context and the model's internal representations just enough to perturb the probability distributions of the surrounding nonsense tokens, thus causing artificial fluctuations in their token-level perplexity, which effectively inflates the total $\Delta$PPL denominator and shifts the median baseline away from zero.}

\begin{figure*}[t]
    \centering
    \includegraphics[width=\textwidth]{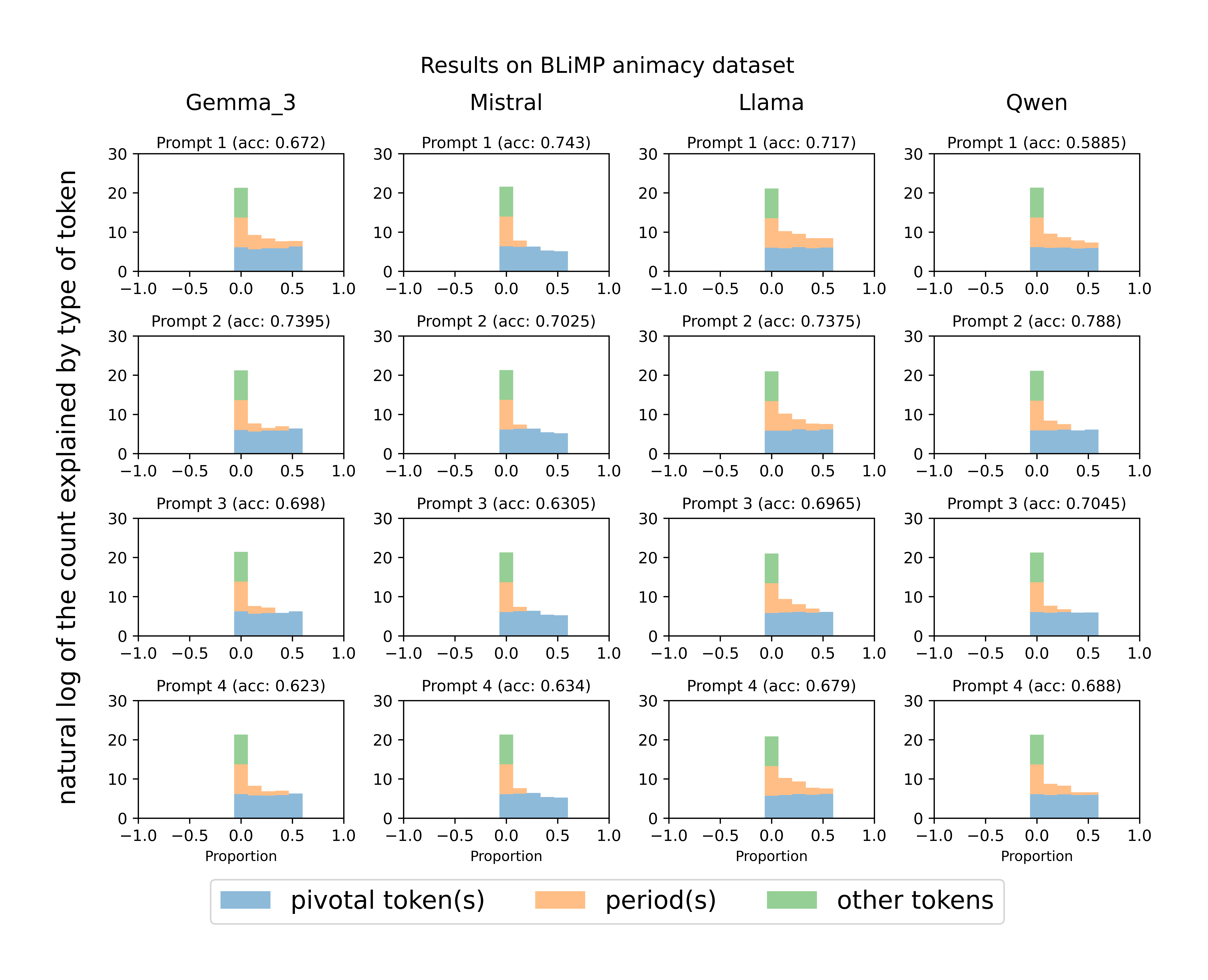}
    \caption{Natural logarithm of the number of sentences (y-axis) in the BLiMP animacy task for which a proportion of the difference in perplexity is explained by key tokens (x-axis). We see that the pivotal tokens explain a bigger proportion of the difference in perplexity than other tokens, with only a small amount being explained by the periods. However, we can notice as also for this task the pivotal tokens are accountable for only a maximum of $\pm 50\%$ of the difference in perplexity.}
    \label{fig:blimp_animacy_results}
\end{figure*}

\begin{figure*}[t]
    \centering
    \includegraphics[width=\textwidth]{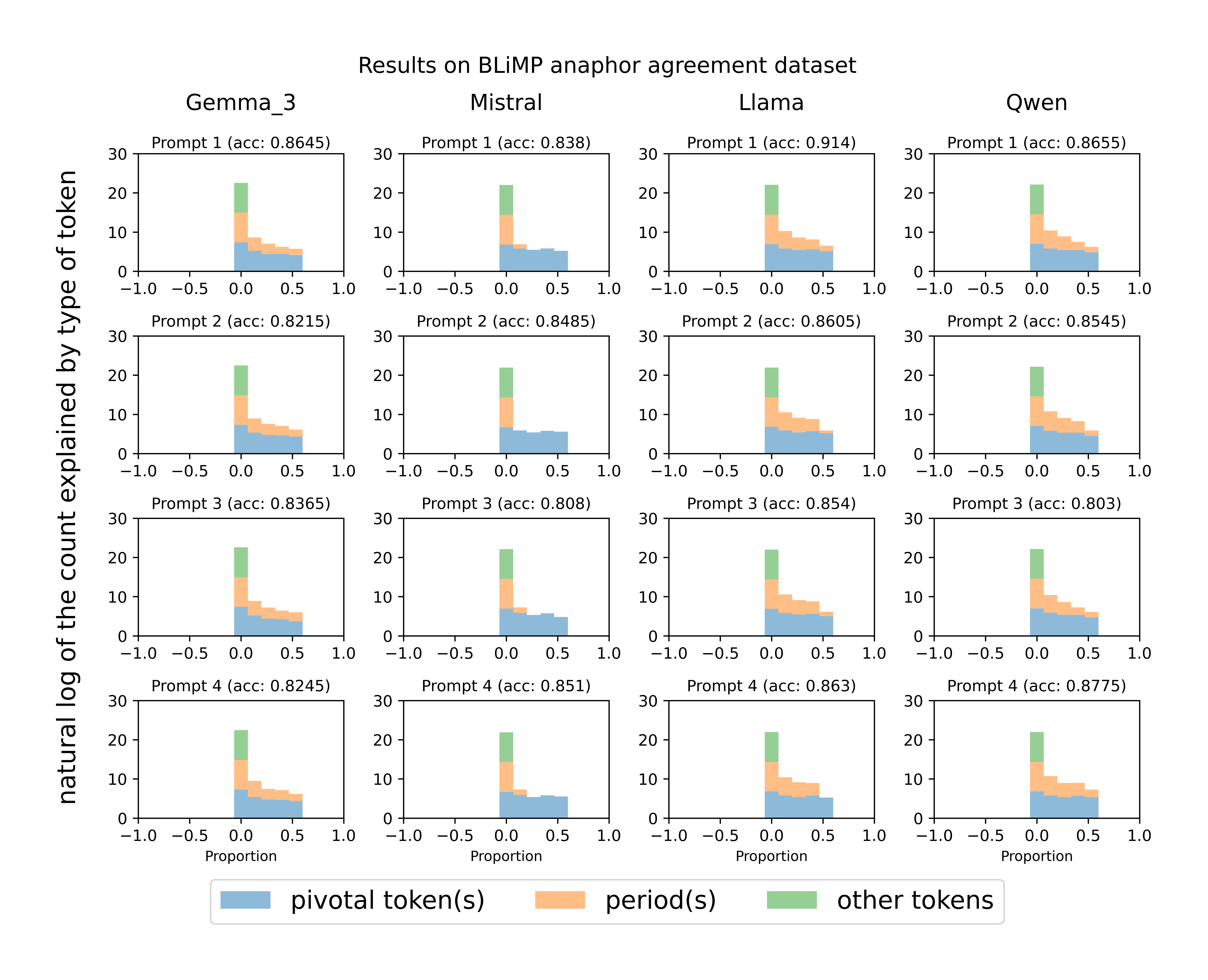}
    \caption{Natural logarithm of the number of sentences (y-axis) in the BLiMP anaphor agreement task for which a proportion of the difference in perplexity is explained by key tokens (x-axis). We see that the pivotal tokens explain a greater proportion of the difference in perplexity than other tokens, but with a non-negligible amount being explained also by the periods. As for the previous task we can notice how the pivotal tokens account for only a maximum of $\pm 50\%$ of the difference in perplexity.}
    \label{fig:blimp_anaphor_results}
\end{figure*}

\begin{figure*}[t]
    \centering
    \includegraphics[width=\textwidth]{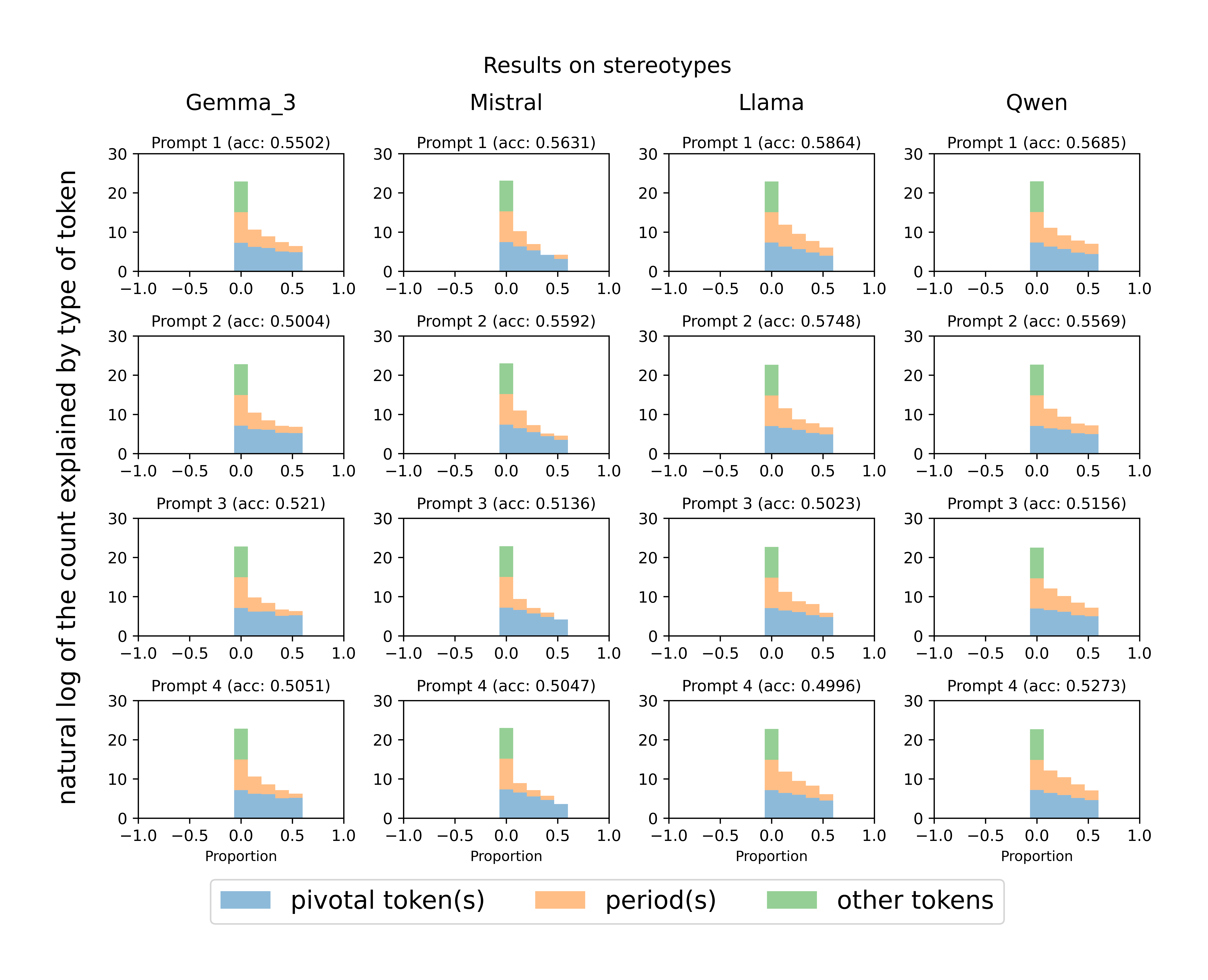}
    \caption{Natural logarithm of the number of sentences (y-axis) in the stereotypes task for which a proportion of the difference in perplexity is explained by key tokens (x-axis). Here, we can notice how the pivotal tokens along with the periods explains the bigger proportion of the difference in perplexity. Again, as for the previous tasks these tokens accounts for only a maximum of $\pm 50\%$ of the difference in perplexity.}
    \label{fig:stereotypes_results}
\end{figure*}

\begin{figure*}
    \centering
    \includegraphics[width=\textwidth]{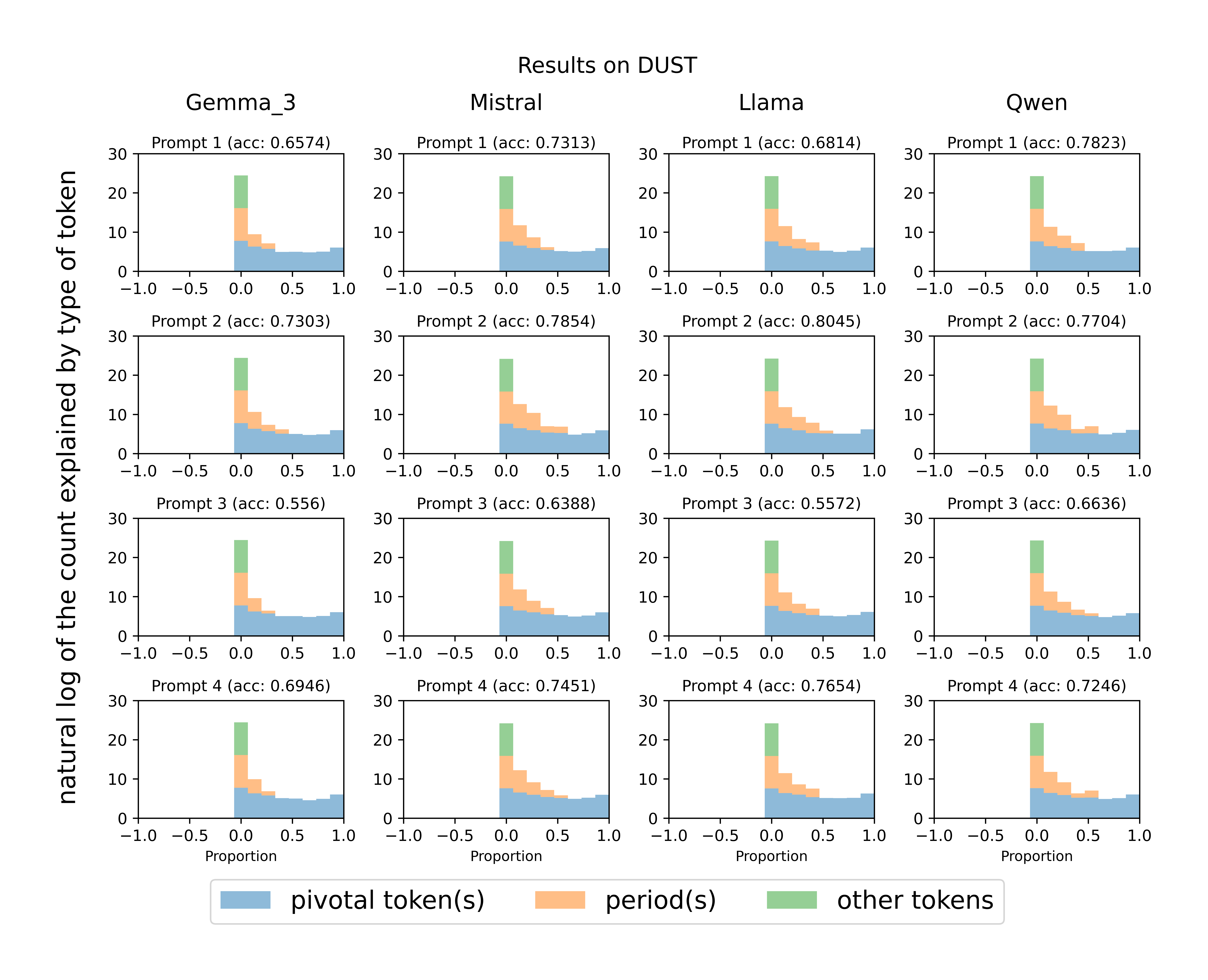}
    \caption{Natural logarithm of the number of sentences (y-axis) in the DUST task for which a proportion of the difference in perplexity is explained by key tokens (x-axis). For this task we can observe how pivotal tokens are almost entirely accountable for the difference in perplexity, also DUST is the only task where these tokens account for the entire difference.}
    \label{fig:dust_results}
\end{figure*}

\end{document}